\makeatletter
\newcommand*{\rom}[1]{\expandafter\@slowromancap\romannumeral #1@}
\makeatother

\documentclass[preprint,twocolumn,5p,12pt]{article}

\usepackage{amssymb}
\usepackage{subcaption}
\usepackage{graphicx}
\usepackage{caption}
\usepackage{xspace}
\usepackage{multirow}
\usepackage{rotating}
 \usepackage{url}
\usepackage{gensymb}
\usepackage [ table ]{ xcolor }
\usepackage[top=1.7cm, bottom=2cm, left=1.5cm, right=1.5cm]{geometry}
\usepackage{setspace}
\usepackage{morefloats}

\usepackage{commath}

\usepackage[affil-it]{authblk}
\usepackage[colorlinks=true,citecolor=blue, urlcolor=blue, linkcolor=blue]{hyperref}

\title{Drastically Reducing the Number of Trainable Parameters in Deep CNNs by Inter-layer Kernel-sharing}

\author{Alireza Azadbakht$ ^{1}$}

\author{Saeed Reza Kheradpisheh$ ^{1,}$\footnote{Corresponding Author \\Email addresses:\\ 
\href{mailto://al.azadbakht@mail.sbu.ac.ir}{al.azadbakht@mail.sbu.ac.ir} (AA), \\\href{mailto://s_kheradpisheh@sbu.ac.ir}{s\_kheradpisheh@sbu.ac.ir} (SRK), \\ \href{mailto://ismail.khalfaoui-hassani@univ-tlse3.fr}{ismail.khalfaoui-hassani@univ-tlse3.fr} (IKH), \\ \href{mailto:/timothee.masquelier@cnrs.fr}{timothee.masquelier@cnrs.fr} (TM)} }

\author{Ismail Khalfaoui-Hassani$ ^{2}$}
\author{Timoth\'ee Masquelier$ ^{3}$}
\affil{\footnotesize $^{1}$ Department of Computer Science, Faculty of Mathematical Sciences, Shahid Beheshti University, Tehran, Iran\\ \footnotesize $^{2} $ Artificial and Natural Intelligence Toulouse Institute (ANITI), Toulouse, France\\\footnotesize $^{3} $  CerCo UMR 5549, CNRS Universit\'e Toulouse 3, Toulouse, France}

\date{}

\begin{document}
\maketitle

\begin{abstract}
Deep convolutional neural networks (DCNNs) have become the state-of-the-art (SOTA) approach for many computer vision tasks: image classification, object detection, semantic segmentation, etc. However, most SOTA networks are too large for edge computing. Here, we suggest a simple way to reduce the number of trainable parameters and thus the memory footprint: sharing kernels between multiple convolutional layers. Kernel-sharing is only possible between ``isomorphic" layers, i.e. layers having the same kernel size, input and output channels. This is typically the case inside each stage of a DCNN. Our experiments on CIFAR-10 and CIFAR-100, using the ConvMixer and SE-ResNet architectures show that the number of parameters of these models can drastically be reduced with minimal cost on accuracy. The resulting networks are appealing for certain edge computing applications that are subject to severe memory constraints, and even more interesting if leveraging ``frozen weights" hardware accelerators. Kernel-sharing is also an efficient regularization method, which can reduce overfitting. The codes are publicly available at \url{https://github.com/AlirezaAzadbakht/kernel-sharing}
\end{abstract}


%

\section{Introduction}
Modern deep learning is pushing the boundaries of artificial intelligence with increasingly complex models. These, in turn, come at a skyrocketing cost in terms of data, energy, memory, computing power, and time to train and use them. Although empirical model-scaling laws exist, such as the one proposed in EfficientNet~\cite{tan2019efficientnet}, the optimal architectures found by these laws still need tens of millions of parameters to reach reasonable accuracies on established tasks and benchmarks~\cite{mahajan2018exploring}~\cite{liu2022convnet}~\cite{dosovitskiy2020image}~\cite{jia2021scaling}~\cite{dai2021coatnet}.  As a result, and despite the huge success of large deep models, they are still not easily usable in resource-constrained systems, such as edge devices and embedded systems~\cite{lane2017squeezing}.
 
A first solution for adapting large CNNs to resource-limited systems is to use compact architectures such as MobileNet~\cite{howard2017mobilenets} designed to minimize the number of computational operations and trainable parameters. Other techniques such as parameter quantization~\cite{hubara2017quantized}~\cite{wu2016quantized} and pruning~\cite{yang2017designing}~\cite{molchanov2016pruning} also  reduce the memory footprint and computational demand of large deep models by reducing the model size. 

To drastically reduce the number of trainable parameters,  we propose sharing parameters between the network layers. Literally, the same set of parameters is being used in several layers while it plays a different role at each layer. This helps to have a smaller set of trainable parameters, without down-scaling the network size. In other words, the network can be folded into layers with shared parameters and deepened without increasing the number of parameters.

When used in CNNs, this technique can be used by sharing  kernels between the isomorphic convolutional layers having the same configuration (i.e. kernel size and the number of input and output channels). In traditional CNNs, weight sharing was limited to neurons in the same feature map. With kernel-sharing, a kernel could be duplicated in feature maps at different layers. To update a shared kernel with backpropagation, the gradients should be accumulated across its different feature maps throughout the network. 

With kernel-sharing, the learning algorithm does not search for layer-specific kernels, rather it looks for shared kernels that can play different roles at different depths of the network. This could be considered as a form of regularization on the network complexity. In usual regularization techniques such as L2-norm~\cite{krogh1991simple}, the complexity of over-parameterized networks is controlled by putting pressure on the majority of parameters to get near zero values. While with shared kernels, the number of parameters is significantly reduced and kernels are forced to maximally exploit their learning capacity. 

 We applied the proposed kernel-sharing to two different deep CNN architectures of ConvMixer~\cite{trockman2022patches} and SE-ResNet~\cite{hu2018squeeze} on CIFAR-10 and CIFAR-100 datasets~\cite{cifar_10}. The first layer of ConvMixer  performs a convolution with large kernels and stride, then, it is followed by a cascade of isomorphic  interlaying depthwise and pointwise convolutional layers. The SE-ResNet architecture has several convolutional stages with squeeze-and-excitation modules~\cite{hu2018squeeze} in between, where, each stage is consist of several isomorphic convolutional layers.

In the extreme case of applying kernel-sharing to all the isomorphic layers, the classification accuracy of the ConvMixer model dropped only by 1.8\% and 3.6\% on CIFAR-10 and CIFAR-100 datasets, while the number of its trainable parameters was reduced by 13.4 and 9.6 times with respect to the baseline model. Similarly, the accuracy drop in SE-ResNet  with kernel-sharing  was negligible, while, the number of its trainable parameters was drastically fewer. 

Also, our results indicate that kernel-sharing can reduce the challenge of overfitting when we keep the network depth or the number of trainable parameters the same as the baseline model.

 



\section{Method}

\begin{figure}[t]
\centering
    \begin{subfigure}[h]{0.5\textwidth}
    \centering
    \includegraphics[scale=0.75]{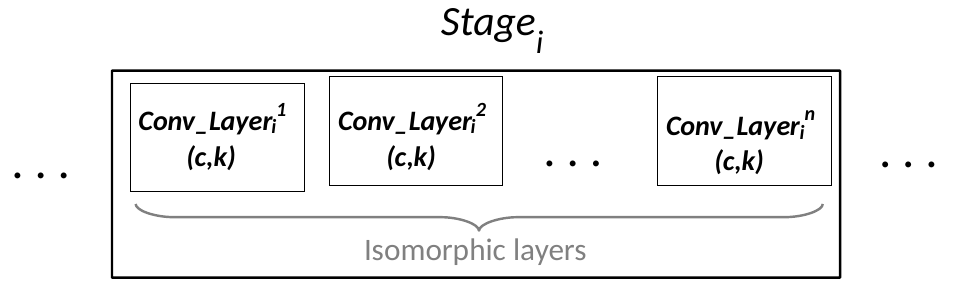}
    \caption{}
    \end{subfigure}
    
    \begin{subfigure}[h]{0.5\textwidth}
    \centering
    \includegraphics[scale=0.75]{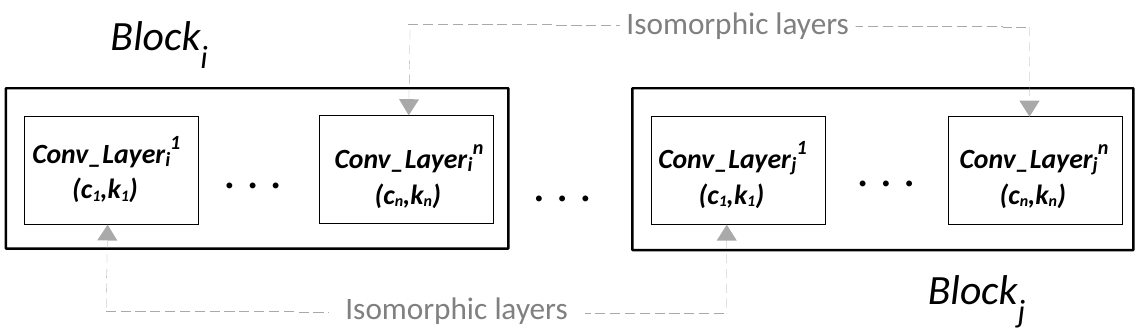}
    \caption{}
    \end{subfigure}
    \caption{ Multi-stage deep CNNs with isomorphic layers have the same number of channels and kernel sizes. a) Isomorphic layers in one stage, where the number of channels and kernel sizes can vary among the stages. b)  Isomorphic layers in multi-block stages, where the number of channels and kernel sizes are different in one block but the same between blocks.}
    \label{fig1}
\end{figure}

State-of-the-art CNNs are usually made of tens of convolutional layers stacked on each other. Usually, some of these layers have similar configurations, such as the same number of channels and kernel size. In multi-stage deep CNNs, usually, the model has several stages of consecutive convolutional layers with the same configuration (see Fig.~\ref{fig1}a). In other words, layer configurations are similar in each stage and vary across stages. For example, the ConvMixer model~\cite{trockman2022patches} has only one stage of convolutional layers, and the SE-ResNet~\cite{hu2018squeeze} model has three or four stages depending on the given dataset. 
Even in deep CNNs with multi-block stages, layers may have different configurations within the same block but they are similar to layers in other blocks (see Fig.~\ref{fig1}b). 

\par
\begin{figure}[t]
    \centering
    \includegraphics[scale=0.7]{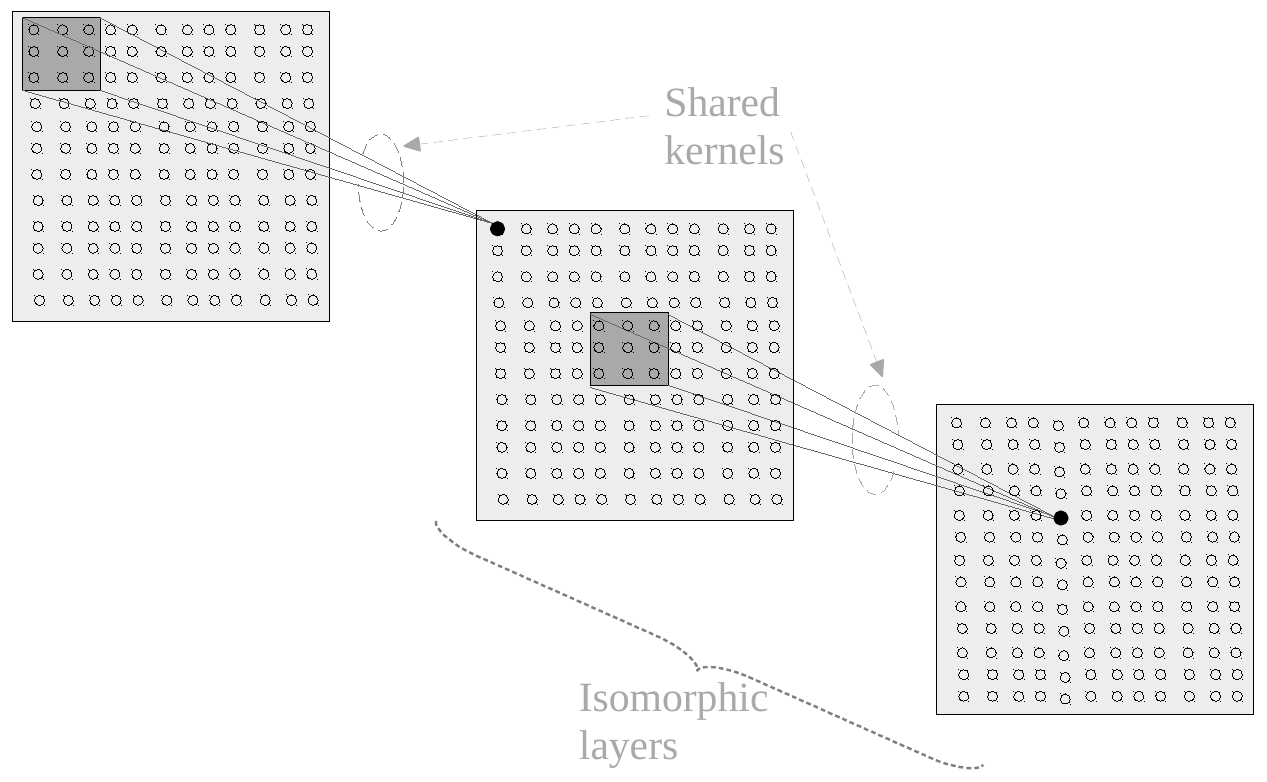}
    \caption{Two consecutive isomorphic layers with shared kernels. The same set of weights is being used in these isomorphic layers. }
    \label{fig2}
\end{figure}

We introduce kernel-sharing or inter-layer weight-sharing for deep CNNs to reduce the number of trainable parameters and eventually reduce the memory footprint, which is especially useful in memory-constrained situations. In the context of CNNs, weight-sharing is an intra-layer concept and refers to using the same kernel at different locations. However, these kernels differ between the layers. Here, we propose to go beyond and train deep CNNs with shared kernels (i.e., kernel weights) among the layers with the same configuration, which we are going to call ``isomorphic layers" hereafter (see Fig.~\ref{fig2}).

\par

We define a sharing group as a set of isomorphic layers sharing their kernels, located in the same (Fig.~\ref{fig1}a) or different stages (Fig.~\ref{fig1}b). Also, one might partition a set of isomorphic layers into two or even more sharing groups. For example, $n$ isomorphic layers in Fig.~\ref{fig1}a could be divided into two sharing groups of $n/2$ isomorphic layers. 

During the forward pass, isomorphic layers in a sharing group use the same set of trainable parameters (i.e., shared kernels). Hence, we should keep these parameters the same after the weight updates following the backward pass. During the training, each layer computes its gradients. Therefore, similar to what happens for the shared weights in CNNs, for each shared kernel, we should accumulate its gradients over its feature maps in all the isomorphic layers of its sharing group.

More formally: let  $w^g$ be the weights of a shared kernel between $n$ isomorphic layers ($l=1,2,...,n$) of the sharing group $g$. Its gradient is:

\begin{equation}
    \frac{\partial L}{\partial w^{g}} =  \sum^n_{l=1} \frac{\partial L}{\partial w^{g}_l},
\end{equation}
where $w^{g}_l$ denotes the use of the shared kernel $w^g$ in the $l^{th}$ isomorphic layer, and $L$ is the loss function of the neural network. 

\par 
Each isomorphic layer in a sharing group computes its own gradients with respect to the shared kernels. Therefore, the computational cost of error-backpropagation in training CNNs with shared kernels is the same as in conventional CNNs. However, the number of trainable parameters is drastically lower with shared kernels. 
\par
The proposed method can be applied to most modern deep learning models by sharing kernels among groups of isomorphic layers. There are no constraints on how many isomorphic layers lie in a sharing group. For example, in a CNN with twenty isomorphic layers, we can have two sharing groups, each containing ten isomorphic layers, or four sharing groups, each containing five isomorphic layers. Kernel-sharing can be even more flexible when applying it to some groups and not others. For example, in the CNN mentioned above, with two groups of ten isomorphic layers, one can apply kernel-sharing to the first group but not the second group. If kernel-sharing applies only to a small part of the network or to a small sharing group, the overall set of trainable parameters will increase. Hence, we have a trade-off here, the more kernel-sharing you have, the fewer trainable parameters you will deal with.

\section{Experimental Results}
Here, we use CIFAR-10 and CIFAR-100 datasets~\cite{cifar_10}, which are benchmark image classification tasks widely used in evaluating deep learning models. The difference between these two datasets is the number of classes, as the name implicates CIFAR-10 contains ten classes, and CIFAR-100 consists of  100 classes. In the following, we evaluate kernel sharing in two recent deep CNN architectures (ConvMixer~\cite{trockman2022patches} and SE-ResNet~\cite{hu2018squeeze}) on these two datasets.

\subsection{Implementation Details}
\par
The ConvMixer model~\cite{trockman2022patches} was initially proposed to investigate if the outstanding performance of Visual Transformer models~\cite{dosovitskiy2020image} was due to their inherently more powerful architecture or because of using patches as an early representation. The model is straightforward: it directly works on patches from input, creating a patch embedding, separating the mixing of spatial and channel dimensions (using depthwise and pointwise convolutions), and maintaining equal channel number and resolution throughout the network. ConvMixer is a single-stage deep CNN consisting of consecutive CNN blocks, each made of a depthwise layer followed by a pointwise layer. The two main hyper-parameters are the number of CNN blocks and the number of channels per block. 
\par
In our experiments, we shared kernels among each of the two sharing groups of the depthwise and pointwise convolutional layers. For example, ConvMixer-384/20 is a model with 20 CNN blocks in depth and 384 channels in each depthwise and pointwise layer, respectively. Therefore, we have two sharing groups, one consists of 20 depthwise layers, and the other one consists of 20 pointwise layers. If we apply kernel-sharing for these two sharing groups, we call this full sharing or two-kernel experiment (see Fig.~\ref{fig3}). We can reduce the granularity of kernel-sharing and split these 20 blocks in half and form four sharing groups, two depthwise and two pointwise sharing groups, each of size ten. We call it the four-kernel experiment. If we go further and split the CNN blocks into four parts, we would have eight sharing groups, called the eight-kernel experiment.
\par
SE-ResNet~\cite{hu2018squeeze} is a variant of ResNet~\cite{he2016deep} that employs squeeze-and-excitation (SE) blocks~\cite{hu2018squeeze} in order to perform dynamic channel-wise feature recalibration. SE-ResNet is a multi-stage ResNet with a SE module at the end of each block of convolutional layers within a stage, which computes a weighted representation of its input channels. The corresponding weight of each channel is a scaling factor computed by a series of fully connected layers.
\par
Depending on the dataset, the SE-ResNet model could have a different number of stages, where each stage consists of a stack of isomorphic convolutional layers. The SE-ResNet models for both CIFAR-10 and CIFAR-100 consist of three stages. We set the convolutional layers in these three stages to have 64, 128, and 256 channels, respectively. 
\par
To demonstrate the effectiveness of kernel-sharing with a stage-wise setting in SE-ResNet experiments, we consider isomorphic layers in each stage as a sharing group. For example, SE-ResNet-d6 has three sharing groups, each consisting of six isomorphic convolutional layers. We investigate kernel-sharing in all the three stages (i.e. sharing groups). We call this experiment as full or all-stages kernel-sharing. In the second experiment, kernel-sharing was applied only to the second and third stages. Then in the third experiment, we only applied kernel-sharing to the third stage and did not enforce kernel-sharing to other stages.


\subsection{ConvMixer Model}

We trained ConvMixer models on both CIFAR datasets using AdamW optimizer\cite{loshchilov2017decoupled} with Cosine learning rate scheduler~\cite{loshchilov2016sgdr} for 200 epochs. Experiments were conducted in batches of 128 images with a learning rate of 0.01.

\begin{figure}[t]
\centering
    \begin{subfigure}[h]{0.50\textwidth}
    \centering
    \includegraphics[scale=0.4]{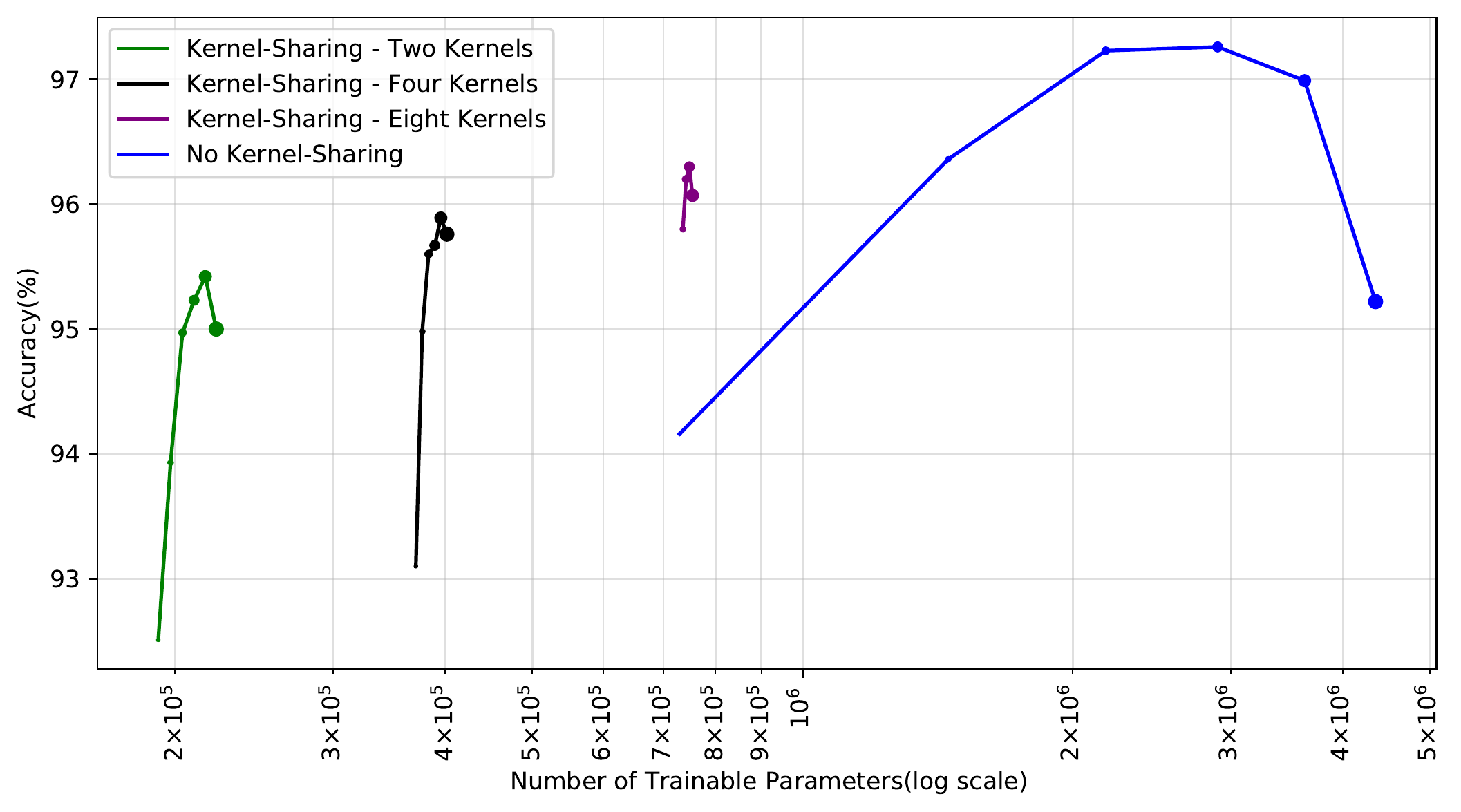}
    \caption{CIFAR-10 dataset.}
    \end{subfigure}
    \hfill
    \begin{subfigure}[h]{0.50\textwidth}
    \centering
    \includegraphics[scale=0.4]{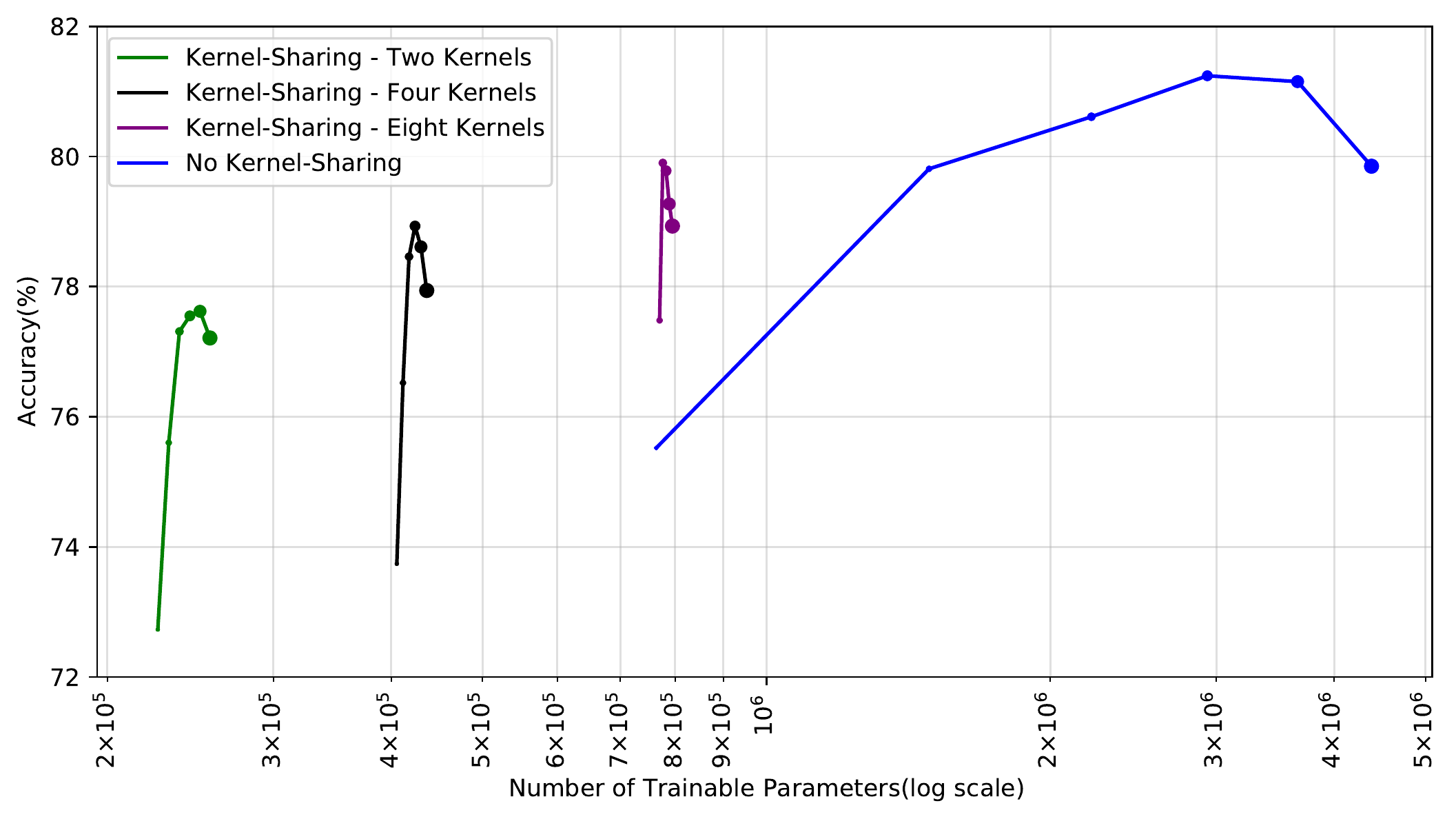}
    \caption{CIFAR-100 dataset.}
    \end{subfigure}
    \caption{The results of ConvMixer-384/$N$ models on the CIFAR family datasets. The depth of the network, $N$, is set to 4, 8, 12, 16, 20, and 24. The radius of each dot represents the number of FLOPs required to run that model. As the depth of the model increases, its computational cost also increases. Note that the eight-kernel model with a depth of four is the same as the baseline, so we ignore this model. The computational demand of models varies in the range of 1.4 to 9 GFLOPs.}
    \label{fig3}
\end{figure}

\par
To demonstrate the effect of kernel-sharing on the ConvMixer model, we set the number of channels in all the layers to 384. 
Fig. \ref{fig3}a illustrates the results of the iso-channel experiment, where we fixed the number of channels and varied the depth of the model. As seen, for the CIFAR-10 dataset, the optimum depths for two-kernel, four-kernel, eight-kernel, and baseline are 20, 20, 16, and 16, respectively.
\par 
The overall accuracy of the optimum full kernel-sharing model (i.e., two-kernel model) is 1.8\% lower than the optimum baseline model (95.4\% vs. 97.2\%), however, its trainable parameters are 13.4 times fewer ($\sim$216K vs. $\sim$2.9M). This means that the kernel-sharing model significantly reduces the memory footprint with only a small cost on accuracy.  Note that their numbers of computations are not so different (7.4 GFLOPs vs. 6 GFLOPs).
\par
In the iso-channel study, we investigated the trade-off between the number of sharing kernels and model performance. If we use four kernels in kernel-sharing, its optimum model is 1.3\% lower than the optimum baseline (95.9\% vs. 97.2\%), with 7.3 times fewer parameters ($\sim$395K vs. $\sim$2.9M).
With eight kernels, this margin reduces to 0.9\% (96.3\% vs. 97.2\%) with 3.9 times fewer parameters ($\sim$748K vs. $\sim$2.9M). Evidently, as the number of sharing groups increases from two to eight, the gap between the baseline and kernel-sharing model shrinks in terms of accuracy and memory footprint. Regarding this trade-off, one can choose to use lower memory or get higher accuracy. Of course, this accuracy drop, even in the extreme case of full kernel-sharing, is not more than 1.8\% while the model size is drastically smaller.
\par
The eight-kernel ConvMixer-384/16  performs similarly to the ConvMixer-384/8 baseline. This equality means that if there are no strong constraints on response time or computation power, but the memory footprint is the main challenge, one can employ the kernel-shared model and approximately reduce the number of trainable parameters by 50\% with almost no drop in the accuracy.
\par
As seen in Fig.~\ref{fig3}(b), the iso-channel study on CIFAR-100 follows the same pattern as CIFAR-10. In this dataset, the optimum depths for two-kernel, four-kernel, eight-kernel, and baseline are 20, 16, 12, and 16, respectively. In comparison with the optimum models on CIFAR-10, it is visible that all the kernel-shared models get to optimum performance with a shallower depth. 
\par
We can also observe the trade-off between the number of shared kernels and model performance in the CIFAR-100 dataset. With two kernels, the difference between the optimum kernel-shared model and optimum baseline is 3.6\% (77.6\% vs. 81.2\%) with 11.7 times fewer parameters ($\sim$250K vs. $\sim$2.9M). If we use four kernels, the difference is 2.3\% (78.9\% vs. 81.2\%) with 6.9 times fewer parameters ($\sim$424K vs. $\sim$2.9M), and with eight kernels, this margin reduces to 1.3\% (79.9\% vs. 81.2\%) with 3.8 times fewer number of parameters ($\sim$776K vs. $\sim$2.9M). 
\par 
With kernel-sharing, we can make the model deeper with a negligible increase in the number of trainable parameters. This small increase is caused by the batch normalization layers, whose parameters are not shared. Indeed, we investigated the effect of sharing batch normalization layers parameters, but it also led to a considerable performance drop. Therefore it is not recommended to share batch normalization layers.


\begin{table*}[h!]
\centering
\begin{tabular}{|cccccc|}
\hline
Model & Depth & No. Channels & No. Parameters & GFLOPs & Accuracy(\%) \\ \hline
Baseline & 4 & 128 & 112138 & 0.22 & 90.28 \\
Kernel-Shared & 4 & 283 & 112644 & 0.86 & 91.29 \\ \hline
Baseline & 8 & 128 & 222218 & 0.46 & 94.28 \\
Kernel-Shared & 8 & 410 & 221820 & 3.36 & 94.76 \\ \hline
\end{tabular}
\caption{The iso-parameter study of ConvMixer model on CIFAR-10 dataset with full kernel-sharing(i.e. two-kernels)}
\label{iso_param_convm}
\end{table*}

\par
The kernel-sharing model drastically decreases the number of parameters compared to the baseline model. In the next experiment we performed a iso-parameter study, in which we compared kernel-shared models with the same number of parameters as the baseline model. To do so, we chose two baselines (ConvMixer-128/4 and ConvMixer-128/8) and compared them with kernel-shared ConvMixer models, which have almost the same number of parameters as the baselines on the CIFAR-10 dataset. Note that we fixed the depth of the model and increased the number of channels so that overall trainable parameters remain in the same order as the baseline (see Table.~\ref{iso_param_convm}).
\par
The kernel-shared ConvMixer-283/4 approximately has the same number of parameters as the baseline ConvMixer-128/4 with 1\% better accuracy on the CIFAR-10 dataset, although with more FLOPs. Also, for baseline ConvMixer-128/8 and kernel-shared ConvMixer-410/8, which have almost the same number of parameters, the kernel-shared model outperforms the baseline by a margin of 0.5\%.

\begin{figure}[t]
\centering
    \begin{subfigure}[h]{0.5\textwidth}
    \centering
    \includegraphics[scale=0.4]{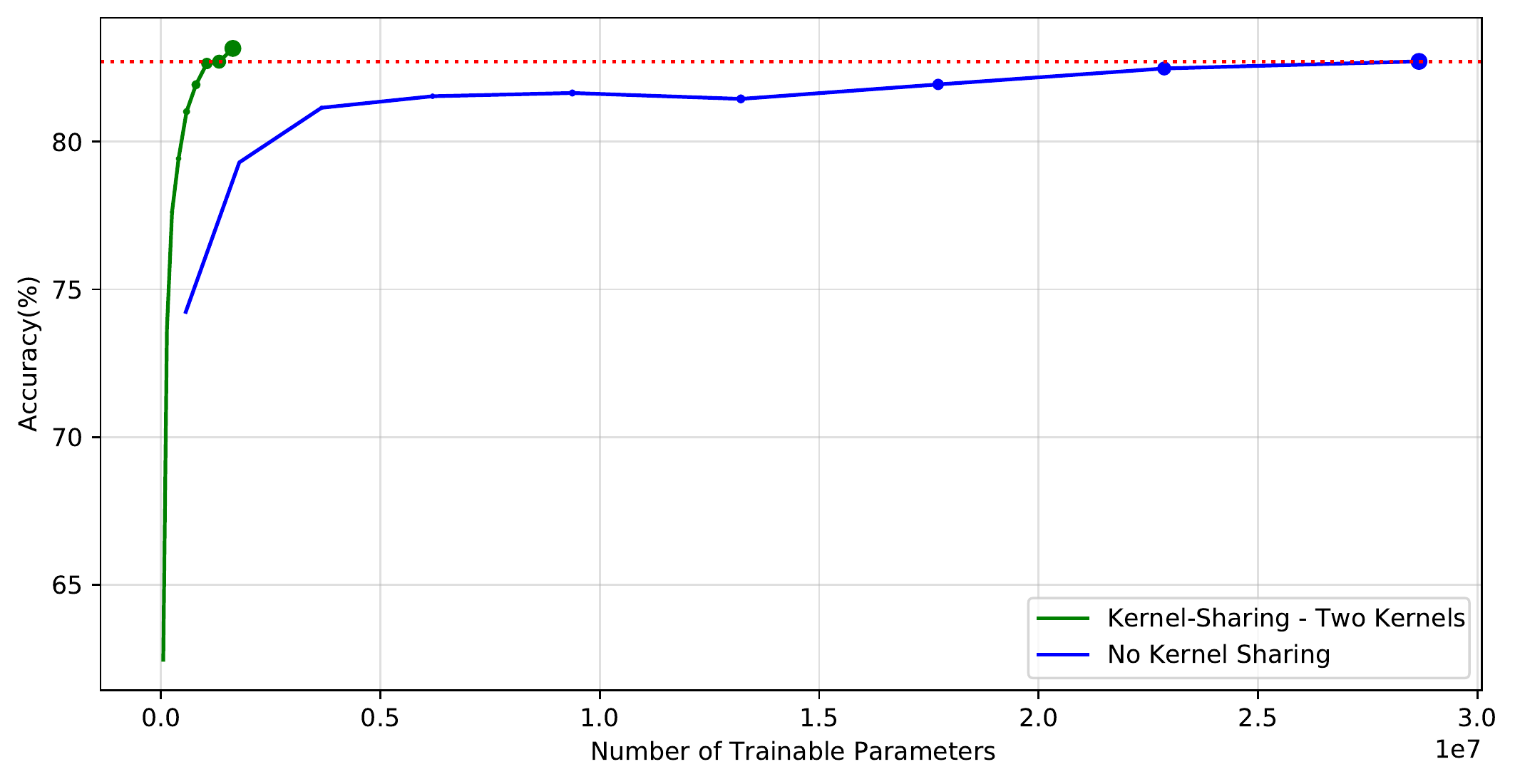}
    \caption{}
    \end{subfigure}
    
    \begin{subfigure}[h]{0.5\textwidth}
    \centering
    \includegraphics[scale=0.4]{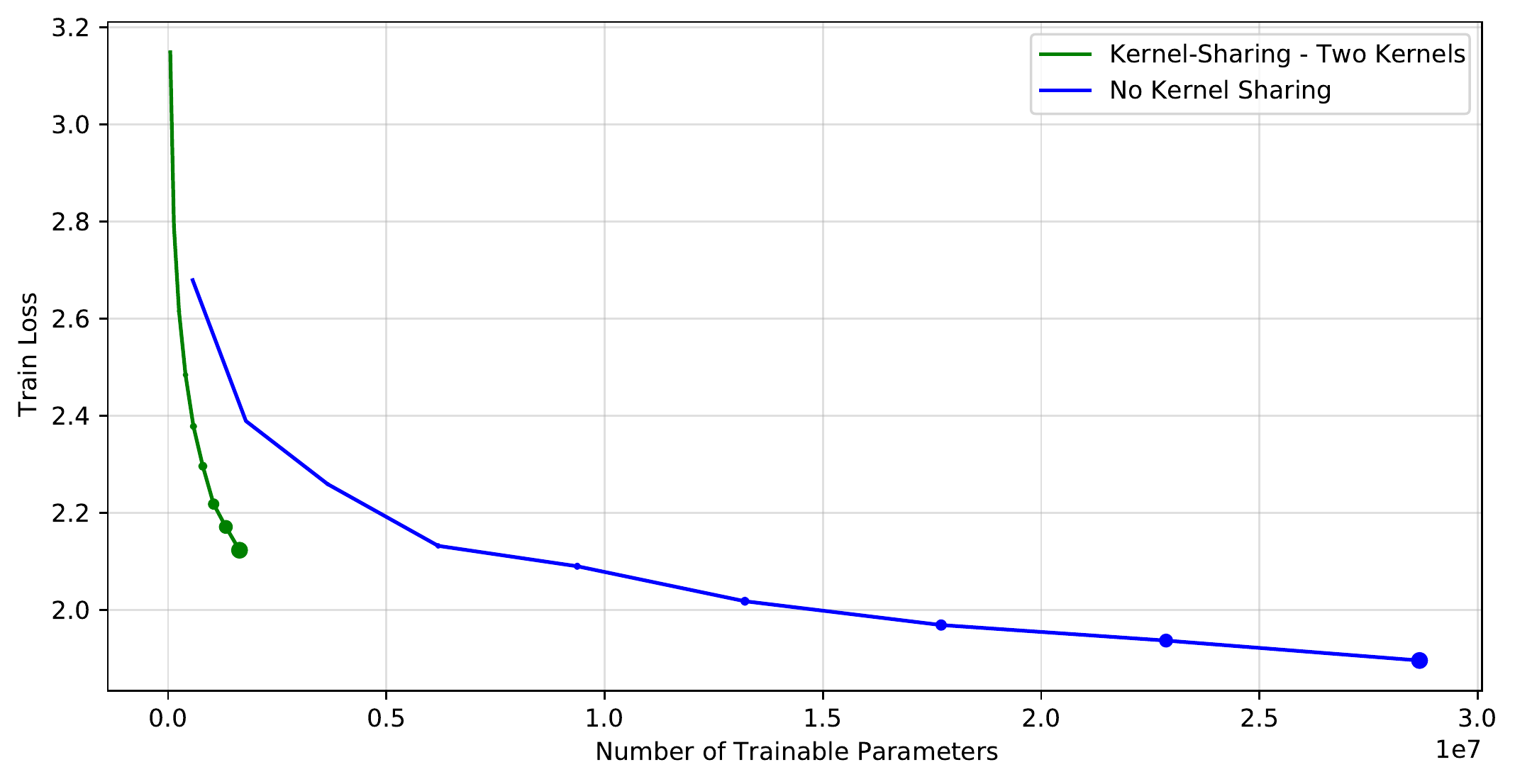}
    \caption{}
    \end{subfigure}
    \caption{Result of the iso-depth study on ConvMixer-$C$/20 models over the CIFAR-100 dataset, where $C$ is 128, 256, 384, 512, 640, 768, 896, 1024, 1152. The radius of each dot represents the number of FLOPs requied to run that model. Obviously, the model computational cost increases with the number of channels. The accuracy of the best baseline model is shown by the red line to make a better comparison. The computational demand of models varies in the range of 1.2 to 58.4 GFLOPs.}
    \label{iso_depth}
\end{figure}
\par

\par
In the iso-depth study, we chose ConvMixer models with the depth fixed at 20 and varied the number of channels from 128 to 1152 with steps of 128. We trained these models on the CIFAR-100 dataset with and without applying kernel-sharing. As shown in Fig.~\ref{iso_depth}a, from ConvMixer-768/20 onwards, kernel-shared models outperform baselines in terms of accuracy. Notably, ConvMixer-768/20 has 16.6 times fewer trainable parameters than the baseline.  
Any increase in channels from 768 would cause the baseline models to overfit. However, kernel-sharing can help the model to avoid overfitting (see Fig.~\ref{iso_depth}b). 
It seems that kernel-sharing can be considered a regularization method, as it reduces the model complexity and increases its generalization.

\subsection{SE-ResNet Model}

Experiments of SE-ResNet were conducted with SGD optimizer and MultiStepLR learning rate scheduler with batches of 128 for 256 epochs and a learning rate of 0.1.

\begin{figure}[t]
\centering
    \begin{subfigure}[h]{0.50\textwidth}
    \centering
    \includegraphics[scale=0.4]{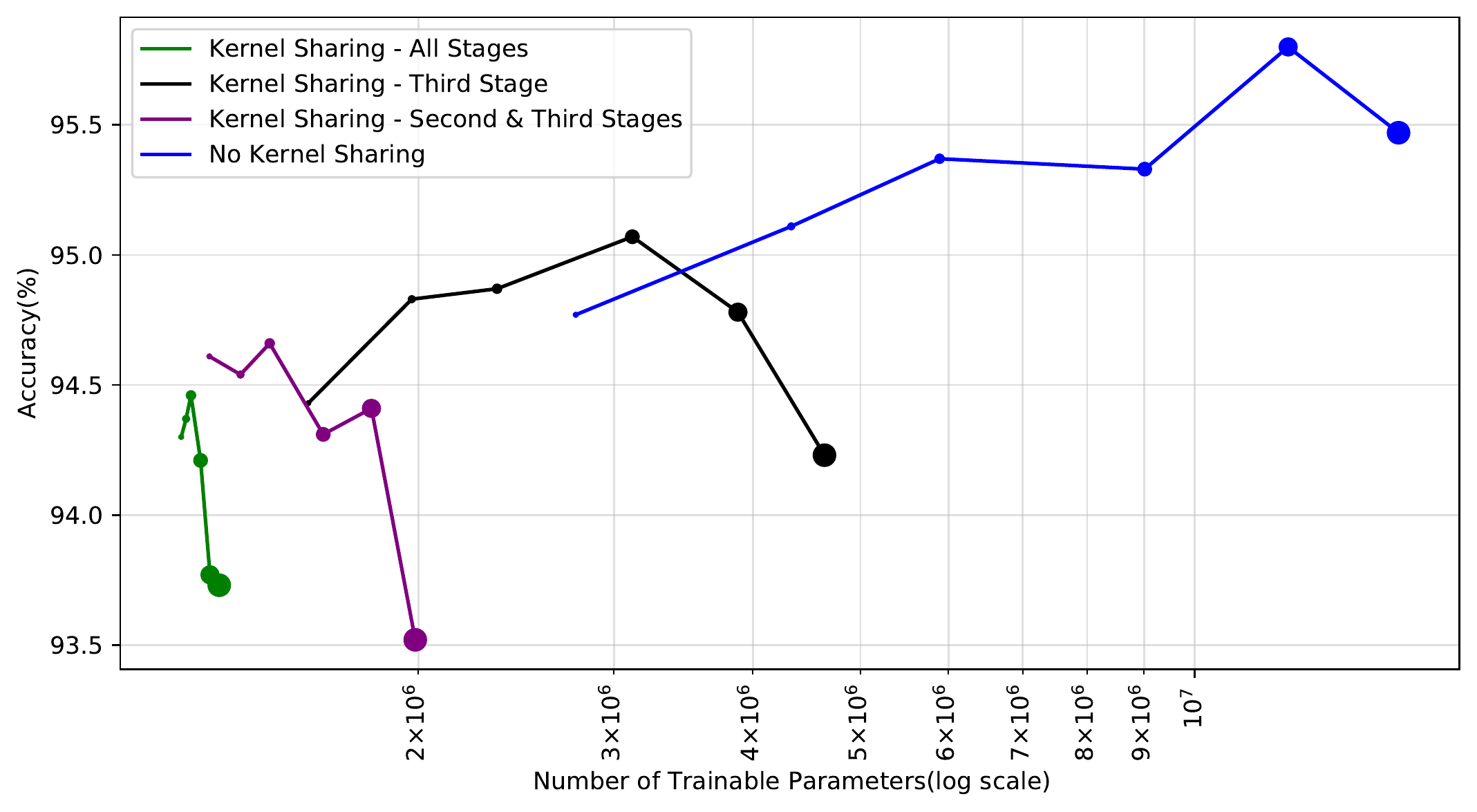}
    \caption{CIFAR-10 dataset.}
    \end{subfigure}
    \hfill
    \begin{subfigure}[h]{0.50\textwidth}
    \centering
    \includegraphics[scale=0.4]{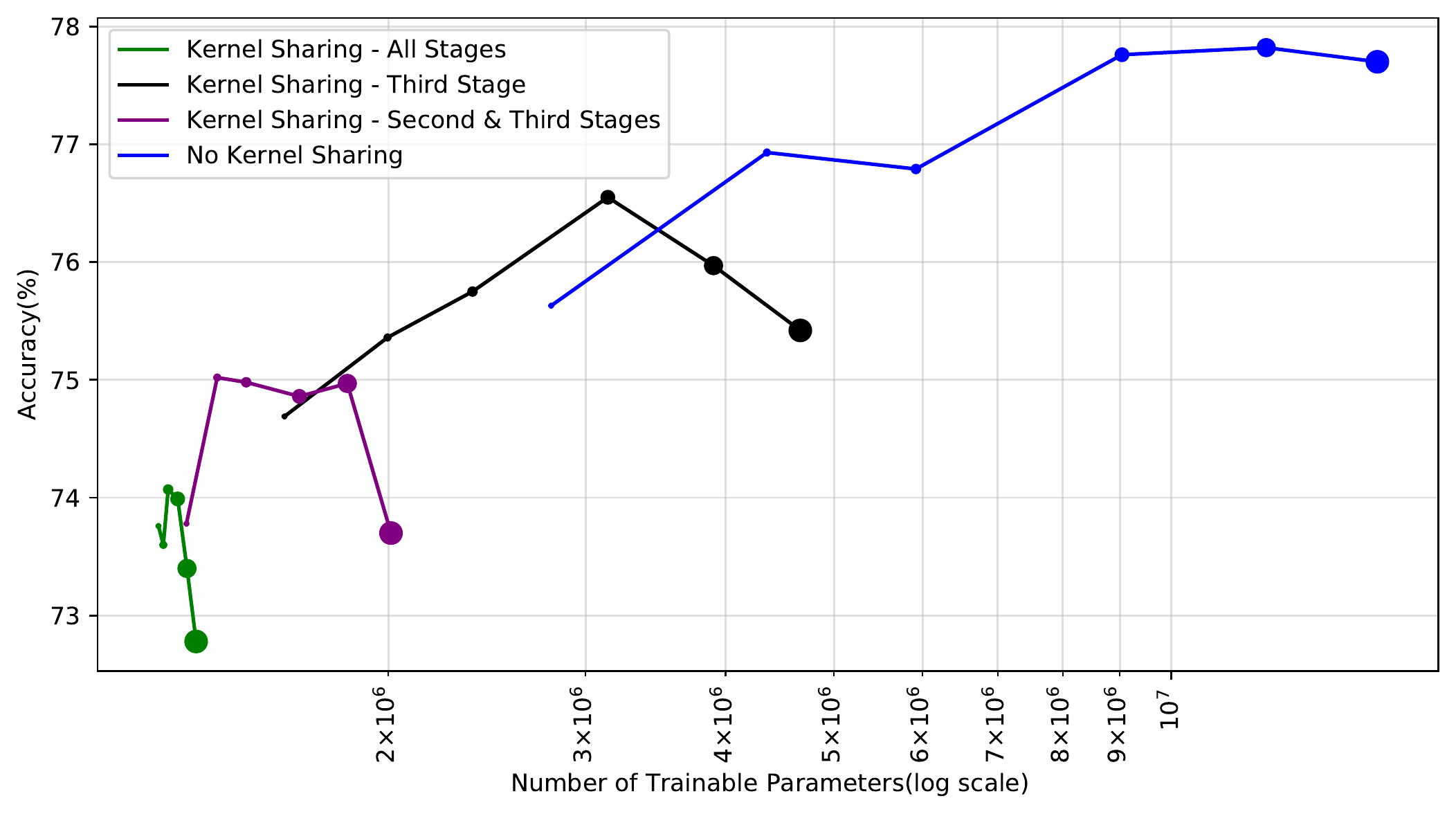}
    \caption{CIFAR-100 dataset.}
    \end{subfigure}
    \caption{Result of SE-ResNet models on the CIFAR family datasets, where the depth of each stage is set to 2, 3, 4, 6, 8, and 10. The radius of each dot represents the number of FLOPs required to run that model. The computational demand of models varies in the range of 0.8 to 4.4 GFLOPs.}
    \label{fig4}
\end{figure}

In the original SE paper\cite{hu2018squeeze}, SE-ResNet-110, and SE-ResNet-164 were used for CIFAR family datasets. These two models consist of 18 and 27 convolutional layers at each stage, so the overall networks have 54 and 81 layers in depth, respectively. 
\par
Here, we decreased the depth of each stage and instead increased the number of channels in our experiments. Indeed, we set the number of channels in the three stages of our SE-ResNet models respectively to 64, 128, and 256 and varied the depth of each stage from two to ten.
Regarding the use of more channels in higher stages, the overall number of trainable parameters increases throughout the network. Therefore, most of the trainable parameters lie in deeper stages. For the iso-channel study, we only investigate the effect of sharing kernels on specific stages. In the SE-ResNet model, each stage could be considered as a sharing group. As the first stage consists of the least number of parameters, sharing kernels only in the first stage is not that beneficial. Therefore, we performed three kernel-sharing experiments: sharing in all stages, sharing in the second and third stages, and sharing only in the third stage.
\par
As seen in Fig.~\ref{fig4}a, on the CIFAR-10 dataset, the optimum baseline is SE-ResNet-d8, with eight isomorphic layers in each stage. SE-ResNet-d6 is the best model when we apply kernel sharing only to the third stage.
The optimum model for applying kernel sharing to all stages or only the second and third stages is SE-ResNet-d4. Seemingly with applying kernel-sharing in more stages, the optimum model is shallower.
\par
By applying kernel-sharing over all stages, the optimum accuracy drops by 1.3\% with respect to the baseline (94.5\% vs. 95.8\%), but with a 9.7 times fewer number of trainable parameters ($\sim$1.2M vs. $\sim$12.1M) and about 0.5 times fewer FLOPs. 
A comparison between kernel-sharing in the second and third stages and baseline reveals that accuracy drops by 1.1\%  (94.7\% vs. 95.8\%) with 8.3 times fewer trainable parameters ($\sim$1.4M vs. $\sim$12.1M) and 0.35 times fewer FLOPs.
The same comparison for kernel-sharing in the third stage has similar results, Where, accuracy drops by 0.7\% (95.1\% vs. 95.8\%) with 3.9 times fewer trainable parameters ($\sim$3.1M vs. $\sim$12.1M) and again with 0.26 times fewer FLOPs. 
The accuracy gap between models with kernel-sharing in all stages and only the second and third stages is negligible. Indeed, when we excluded the first stage from kernel-sharing, the number of trainable parameters remained approximately the same as in the all-stages kernel-sharing model. However, we gained a slightly (0.2\%) better accuracy.
\par 
It can be seen that, as for the ComvMixer models, when we apply kernel-sharing to more stages, the accuracy gap between the kernel-sharing and baseline SE-ResNet models slightly increases with 1.3\% in the most extreme case. On the other hand, the number of parameters of kernel-shared models drastically decreases to 1.2M from 12.1M.
\par
The SE-ResNet-d6 with kernel-sharing only in the third stage has the same accuracy as the SE-ResNet-d3 baseline. Therefore, one can employ the kernel-shared model and save approximately 28\% trainable parameters without any drop in accuracy.
\par
On the CIFAR-100 dataset, the optimum depth for the baseline is eight (SE-ResNet-d8). For all-stages kernel-sharing, the optimum depth is half the baseline (SE-ResNet-d4). When we apply kernel-sharing to the second and third stages, the optimum depth is three (SE-ResNet-d3), and for kernel sharing only in the third stage, the optimum depth is six (SE-ResNet-d6).
\par
The iso-channel study on the CIFAR-100 dataset demonstrates that after applying kernel-sharing over all stages, the overall accuracy drops by 3.7\% (from 77.8\% to 74.1\%) but with 9.6 times smaller number of trainable parameters ($\sim$12.1M vs. $\sim$1.3M) and with about half the FLOPs. 
 Comparison between baseline and kernel-sharing in the second and third stages reveals that overall accuracy drops by 2.8\%  (77.8\% vs. 75\%) with 8.64 times smaller number of trainable parameters ($\sim$12.1M vs. $\sim$1.4M) and fewer FLOPs.
The same comparison for kernel-sharing only in the third stage has similar results. The overall accuracy drops by 1.3\% (77.8\% vs. 76.5\%) with a 3.9 times smaller number of trainable parameters ($\sim$12.1M vs. $\sim$3.1M) and still efficient in terms of required FLOPs.

\section{Discussion}
In this paper, we proposed inter-layer kernel-sharing for convolutional deep neural networks. This new method helps to reduce the number of parameters in deep CNNs with a slight drop in accuracy. This idea is a generalization of weight sharing in conventional CNNs. Traditionally in CNNs, weights are shared among neurons in a feature map, but here we pushed it further and applied weight sharing between the isomorphic layers. In this way, isomorphic layers in a sharing group share the same set of kernels. With kernel-sharing, a single layer is not free to learn a set of kernels that would be only effective in that specific layer. Instead, isomorphic layers in a sharing group should find a more general set of kernels that would be effective for all the layers in that group.
\par
\par
It seems that kernel-sharing could be thought of as a particular case of unfolded recurrent neural networks, where at each time step, there is no input token, and the model only does its computation on the feedback of the previous time step, and the output is the final state.
\par
As seen in the iso-parameter study on ConvMixer models, we can conclude that kernel-sharing on shallower networks is more effective. This phenomenon could be justified by the fact that when the depth of the model increases, the kernel-shared model faces a more challenging problem in finding the best kernel that would work if used repeatedly in many layers. When the depth increases, too many gradient updates apply on the same kernel, and some of the effects smooth out, which negatively affects the overall performance.
\par
In hardware implementations, a small non-programmable circuit can be designed to do the computation needed for a neural network~\cite{barry2015always}~\cite{reagen2016minerva}~\cite{buckler2018eva2}. This hardwired model implementation is more efficient and faster than the same model on a programable circuit~\cite{whatmough2019fixynn}~\cite{samajdar2018scale}. Many optimizations like fixed shift-add scalers~\cite{10.1145/504709.504710}, zero-overhead weight pruning~\cite{parashar2017scnn}, optimized intermediate precision~\cite{whatmough2019fixynn}, zero DRAM bandwidth~\cite{whatmough2019fixynn}~\cite{shih2020zebra}, and minimal activation storage~\cite{whatmough2019fixynn} could be used in these kinds of kits. The hardware implementation of a kernel-shared model is much more efficient and accessible. For a model with a single set of kernels shared among the whole network, one only needs to design a CNN layer on the circuit and use it repeatedly to run the whole network. In this way, the silicon area needed for implementing a neural network drastically decreases, and the memory footprint would be significantly smaller. 
\par
In the iso-parameter case, applying kernel-sharing improves the model's accuracy, but the resulting model has more FLOPs than the baseline counterpart. For example, in terms of accuracy, the kernel-shared ConvMixer-283/4 is better than the ConvMixer-128/4 baseline with the same number of parameters. This behavior of kernel-sharing implies that if computational demand is not an issue, we can use kernel-sharing and achieve better performance. In practice, the benefits of hardware implementation of the kernel-shared model can compensate for its higher computational demand. 
\par
As seen in the iso-depth study with fixed depth, when we increase the width of the network (number of channels), at some point, baseline models start to overfit. However, kernel-shared counterparts outperform them with fewer trainable parameters and the same number of FLOPs. In this way, we can consider kernel-sharing as a regularization method. Therefore, CNN blocks consisting of convolutional layers with shared kernels could be used as the building blocks of large networks with a lower risk of overfitting.

\par
In iso-channel studies on ConvMixer and SE-ResNet models on both CIFAR-10 and CIFAR-100 datasets, we faced a trade-off between the number of trainable parameters and the classification accuracy. If we enforce kernel-sharing more extensively, the model would be drastically smaller in terms of trainable parameters, however, the accuracy would drop a bit. In the worst case, the accuracy dropped by 1.8\% in CIFAR-10 and 3.7\% in CIFAR-100 datasets. Regarding this trade-off, we have to pay attention to our memory constraints or the available silicon area in hardware implementations. We should first fix the number of trainable parameters and the feasible computation power, then, choose the proper depth and the number of shared kernels.
\par
\par
The main question is why we need kernel-sharing and where it would be helpful. Most accurate deep neural networks have a large number of trainable parameters, and they are too heavy to be implemented on a small silicon area of an edge device~\cite{lane2017squeezing}.
Kernel-sharing can be the solution to this problem. We could achieve a model with drastically lower trainable parameters but with a slight drop in accuracy. Note that kernel-sharing is very flexible, and one can choose how and where to apply it. 
\par

\bibliographystyle{IEEEtran}

\begin{thebibliography}{10}
\providecommand{\url}[1]{#1}
\csname url@samestyle\endcsname
\providecommand{\newblock}{\relax}
\providecommand{\bibinfo}[2]{#2}
\providecommand{\BIBentrySTDinterwordspacing}{\spaceskip=0pt\relax}
\providecommand{\BIBentryALTinterwordstretchfactor}{4}
\providecommand{\BIBentryALTinterwordspacing}{\spaceskip=\fontdimen2\font plus
\BIBentryALTinterwordstretchfactor\fontdimen3\font minus
  \fontdimen4\font\relax}
\providecommand{\BIBforeignlanguage}[2]{{%
\expandafter\ifx\csname l@#1\endcsname\relax
\typeout{** WARNING: IEEEtran.bst: No hyphenation pattern has been}%
\typeout{** loaded for the language `#1'. Using the pattern for}%
\typeout{** the default language instead.}%
\else
\language=\csname l@#1\endcsname
\fi
#2}}
\providecommand{\BIBdecl}{\relax}
\BIBdecl

\bibitem{tan2019efficientnet}
M.~Tan and Q.~Le, ``Efficientnet: Rethinking model scaling for convolutional
  neural networks,'' in \emph{International conference on machine
  learning}.\hskip 1em plus 0.5em minus 0.4em\relax PMLR, 2019, pp. 6105--6114.

\bibitem{mahajan2018exploring}
D.~Mahajan, R.~Girshick, V.~Ramanathan, K.~He, M.~Paluri, Y.~Li, A.~Bharambe,
  and L.~Van Der~Maaten, ``Exploring the limits of weakly supervised
  pretraining,'' in \emph{Proceedings of the European conference on computer
  vision (ECCV)}, 2018, pp. 181--196.

\bibitem{liu2022convnet}
Z.~Liu, H.~Mao, C.-Y. Wu, C.~Feichtenhofer, T.~Darrell, and S.~Xie, ``A convnet
  for the 2020s,'' in \emph{Proceedings of the IEEE/CVF Conference on Computer
  Vision and Pattern Recognition}, 2022, pp. 11\,976--11\,986.

\bibitem{dosovitskiy2020image}
A.~Dosovitskiy, L.~Beyer, A.~Kolesnikov, D.~Weissenborn, X.~Zhai,
  T.~Unterthiner, M.~Dehghani, M.~Minderer, G.~Heigold, S.~Gelly \emph{et~al.},
  ``An image is worth 16x16 words: Transformers for image recognition at
  scale,'' \emph{arXiv preprint arXiv:2010.11929}, 2020.

\bibitem{jia2021scaling}
C.~Jia, Y.~Yang, Y.~Xia, Y.-T. Chen, Z.~Parekh, H.~Pham, Q.~Le, Y.-H. Sung,
  Z.~Li, and T.~Duerig, ``Scaling up visual and vision-language representation
  learning with noisy text supervision,'' in \emph{International Conference on
  Machine Learning}.\hskip 1em plus 0.5em minus 0.4em\relax PMLR, 2021, pp.
  4904--4916.

\bibitem{dai2021coatnet}
Z.~Dai, H.~Liu, Q.~V. Le, and M.~Tan, ``Coatnet: Marrying convolution and
  attention for all data sizes,'' \emph{Advances in Neural Information
  Processing Systems}, vol.~34, pp. 3965--3977, 2021.

\bibitem{lane2017squeezing}
N.~D. Lane, S.~Bhattacharya, A.~Mathur, P.~Georgiev, C.~Forlivesi, and
  F.~Kawsar, ``Squeezing deep learning into mobile and embedded devices,''
  \emph{IEEE Pervasive Computing}, vol.~16, no.~3, pp. 82--88, 2017.

\bibitem{howard2017mobilenets}
A.~G. Howard, M.~Zhu, B.~Chen, D.~Kalenichenko, W.~Wang, T.~Weyand,
  M.~Andreetto, and H.~Adam, ``Mobilenets: Efficient convolutional neural
  networks for mobile vision applications,'' \emph{arXiv preprint
  arXiv:1704.04861}, 2017.

\bibitem{hubara2017quantized}
I.~Hubara, M.~Courbariaux, D.~Soudry, R.~El-Yaniv, and Y.~Bengio, ``Quantized
  neural networks: Training neural networks with low precision weights and
  activations,'' \emph{The Journal of Machine Learning Research}, vol.~18,
  no.~1, pp. 6869--6898, 2017.

\bibitem{wu2016quantized}
J.~Wu, C.~Leng, Y.~Wang, Q.~Hu, and J.~Cheng, ``Quantized convolutional neural
  networks for mobile devices,'' in \emph{Proceedings of the IEEE conference on
  computer vision and pattern recognition}, 2016, pp. 4820--4828.

\bibitem{yang2017designing}
T.-J. Yang, Y.-H. Chen, and V.~Sze, ``Designing energy-efficient convolutional
  neural networks using energy-aware pruning,'' in \emph{Proceedings of the
  IEEE conference on computer vision and pattern recognition}, 2017, pp.
  5687--5695.

\bibitem{molchanov2016pruning}
P.~Molchanov, S.~Tyree, T.~Karras, T.~Aila, and J.~Kautz, ``Pruning
  convolutional neural networks for resource efficient inference,'' \emph{arXiv
  preprint arXiv:1611.06440}, 2016.

\bibitem{krogh1991simple}
A.~Krogh and J.~Hertz, ``A simple weight decay can improve generalization,''
  \emph{Advances in neural information processing systems}, vol.~4, 1991.

\bibitem{trockman2022patches}
A.~Trockman and J.~Z. Kolter, ``Patches are all you need?'' \emph{arXiv
  preprint arXiv:2201.09792}, 2022.

\bibitem{hu2018squeeze}
J.~Hu, L.~Shen, and G.~Sun, ``Squeeze-and-excitation networks,'' in
  \emph{Proceedings of the IEEE conference on computer vision and pattern
  recognition}, 2018, pp. 7132--7141.

\bibitem{cifar_10}
A.~Krizhevsky and G.~Hinton, ``Learning multiple layers of features from tiny
  images,'' \emph{(Technical Report) University of Toronto}, 2009.

\bibitem{he2016deep}
K.~He, X.~Zhang, S.~Ren, and J.~Sun, ``Deep residual learning for image
  recognition,'' in \emph{Proceedings of the IEEE conference on computer vision
  and pattern recognition}, 2016, pp. 770--778.

\bibitem{loshchilov2017decoupled}
I.~Loshchilov and F.~Hutter, ``Decoupled weight decay regularization,''
  \emph{arXiv preprint arXiv:1711.05101}, 2017.

\bibitem{loshchilov2016sgdr}
------, ``Sgdr: Stochastic gradient descent with warm restarts,'' \emph{arXiv
  preprint arXiv:1608.03983}, 2016.

\bibitem{barry2015always}
B.~Barry, C.~Brick, F.~Connor, D.~Donohoe, D.~Moloney, R.~Richmond,
  M.~O'Riordan, and V.~Toma, ``Always-on vision processing unit for mobile
  applications,'' \emph{IEEE Micro}, vol.~35, no.~2, pp. 56--66, 2015.

\bibitem{reagen2016minerva}
B.~Reagen, P.~Whatmough, R.~Adolf, S.~Rama, H.~Lee, S.~K. Lee, J.~M.
  Hern{\'a}ndez-Lobato, G.-Y. Wei, and D.~Brooks, ``Minerva: Enabling
  low-power, highly-accurate deep neural network accelerators,'' in \emph{2016
  ACM/IEEE 43rd Annual International Symposium on Computer Architecture
  (ISCA)}.\hskip 1em plus 0.5em minus 0.4em\relax IEEE, 2016, pp. 267--278.

\bibitem{buckler2018eva2}
M.~Buckler, P.~Bedoukian, S.~Jayasuriya, and A.~Sampson, ``Eva$^2$: Exploiting
  temporal redundancy in live computer vision,'' in \emph{2018 ACM/IEEE 45th
  Annual International Symposium on Computer Architecture (ISCA)}.\hskip 1em
  plus 0.5em minus 0.4em\relax IEEE, 2018, pp. 533--546.

\bibitem{whatmough2019fixynn}
P.~N. Whatmough, C.~Zhou, P.~Hansen, S.~K. Venkataramanaiah, J.-s. Seo, and
  M.~Mattina, ``Fixynn: Efficient hardware for mobile computer vision via
  transfer learning,'' \emph{arXiv preprint arXiv:1902.11128}, 2019.

\bibitem{samajdar2018scale}
A.~Samajdar, Y.~Zhu, P.~Whatmough, M.~Mattina, and T.~Krishna, ``Scale-sim:
  Systolic cnn accelerator simulator,'' \emph{arXiv preprint arXiv:1811.02883},
  2018.

\bibitem{10.1145/504709.504710}
K.~D. Cooper, L.~T. Simpson, and C.~A. Vick, ``Operator strength reduction,''
  \emph{ACM Trans. Program. Lang. Syst.}, vol.~23, no.~5, p. 603–625, sep
  2001.

\bibitem{parashar2017scnn}
A.~Parashar, M.~Rhu, A.~Mukkara, A.~Puglielli, R.~Venkatesan, B.~Khailany,
  J.~Emer, S.~W. Keckler, and W.~J. Dally, ``Scnn: An accelerator for
  compressed-sparse convolutional neural networks,'' \emph{ACM SIGARCH computer
  architecture news}, vol.~45, no.~2, pp. 27--40, 2017.

\bibitem{shih2020zebra}
H.-T. Shih and T.-S. Chang, ``Zebra: memory bandwidth reduction for cnn
  accelerators with zero block regularization of activation maps,'' in
  \emph{2020 IEEE International Symposium on Circuits and Systems
  (ISCAS)}.\hskip 1em plus 0.5em minus 0.4em\relax IEEE, 2020, pp. 1--5.

\end{thebibliography}

\end{document}